\documentclass{llncs}
\usepackage{llncsdoc}
\usepackage{times}

\usepackage{graphicx} 
\usepackage{subfigure} 
\usepackage{slashbox}

\usepackage{array}

\usepackage{algorithm}
\usepackage{algorithmic}
\usepackage{cleveref}
\usepackage{hyperref}
\usepackage{amsfonts,mathrsfs}
\usepackage{amsmath,amssymb,graphicx,textcomp,enumerate,bbm,latexsym,xcolor}

\usepackage{times}

\usepackage{graphicx} 
\usepackage{subfigure} 
\usepackage{array}
\usepackage{url}
\usepackage{color}
\usepackage{lipsum}
\usepackage{multirow}
\usepackage{multicol}
\usepackage{float}
\usepackage{mydefs}

\begin{document}

\title{A Simple Exponential Family Framework \\ for Zero-Shot Learning}
\titlerunning{A Simple Exponential Family Framework for Zero-Shot Learning}

\author{Vinay Kumar Verma$^\sharp$ \and Piyush Rai$^\sharp$}    

\authorrunning{Verma,Wang,Rai}

\institute{
$^\sharp$Dept. of Computer Science \& Engineering, IIT Kanpur, India \\
%$^\dagger$ECE Department, Duke University \\
\texttt{\{vkverma,piyush\}@cse.iitk.ac.in}\\ %, \texttt{ww107@duke.edu}\\
}

\maketitle
\vspace{-1em}
\begin{abstract} 
    We present a simple generative framework for learning to predict previously unseen classes, based on estimating \emph{class-attribute-gated} class-conditional distributions. We model each class-conditional distribution as an exponential family distribution and the parameters of the distribution of each seen/unseen class are defined as functions of the respective observed class attributes. These functions can be learned using only the seen class data and can be used to predict the parameters of the class-conditional distribution of each unseen class. Unlike most existing methods for zero-shot learning that represent classes as fixed embeddings in some vector space, our generative model naturally represents each class as a probability distribution. It is simple to implement and also allows leveraging additional unlabeled data from unseen classes to improve the estimates of their class-conditional distributions using transductive/semi-supervised learning. Moreover, it extends seamlessly to few-shot learning by easily updating these distributions when provided with a small number of additional labelled examples from unseen classes. Through a comprehensive set of experiments on several benchmark data sets, we demonstrate the efficacy of our framework\footnote{Code \& data are available:  \url{https://github.com/vkverma01/Zero-Shot/}}. 
\end{abstract} 

\section{Introduction}

The problem of learning to predict unseen classes, also popularly known as Zero-Shot Learning (ZSL), is an important learning paradigm which refers to the problem of recognizing objects from classes that were not seen at training time~\cite{lampert2009learning,socher2013zero}. ZSL is especially relevant for learning ``in-the-wild'' scenarios, where new concepts need to be discovered on-the-fly, without having access to labelled data from the novel classes/concepts. This has led to a tremendous amount of interest in developing ZSL methods that can learn in a robust and scalable manner, even when the amount of supervision for the classes of interest is relatively scarce.

A large body of existing prior work for ZSL is based on embedding the data into a semantic vector space, where distance based methods can be applied to find the most likely class which itself is represented as a point in the same semantic space~\cite{socher2013zero,norouzi2013zero,zhang2015zero56}. However, a limitation of these methods is that each class is represented as a fixed point in the embedding space which does not adequately account for intra-class variability~\cite{akata2015evaluation,mukherjee2016gaussian}. We provide a more detailed overview of existing work on ZSL in the Related Work section.

Another key limitation of most of the existing methods is that they usually lack a proper generative model of the data. Having a generative model has several advantages~\cite{murphy2012machine}. For example, (1) data of different types can be modeled in a principled way using appropriately chosen class-conditional distributions; (2) unlabeled data can be seamlessly integrated (for both seen as well as unseen classes) during parameter estimation, leading to a transductive/semi-supervised estimation procedure, which may be useful when the amount of labeled data for the seen classes is small, or if the distributions of seen and unseen classes are different from each other~\cite{unsupervised2015domain}; and (3) a rich body of work, both frequentist and Bayesian, on learning generative models~\cite{murphy2012machine} can be brought to bear during the ZSL parameter estimation process.

Motivated by these desiderata, we present a generative framework for zero-shot learning. Our framework is based on modelling the class-conditional distributions of seen as well as unseen classes using exponential family distributions~\cite{brown1986fundamentals}, and further conditioning the parameters of these distributions on the respective class-attribute vectors via a linear/nonlinear regression model of one's choice. The regression model allows us to predict the parameters of the class-conditional distributions of \emph{unseen} classes using only their class attributes, enabling us to perform zero-shot learning.

In addition to the generality and modelling flexibility of our framework, another of its appealing aspects is its simplicity. In contrast with various other state-of-the-art methods, our framework is very simple to implement and easy to extend. In particular, as we will show, parameter estimation in our framework simply reduces to solving a linear/nonlinear regression problem, for which a closed-form solution exists. Moreover, extending our framework to incorporate unlabeled data from the unseen classes, or a small number of labelled examples from the unseen classes, i.e., performing few-shot learning~\cite{salakhutdinov2013learning,mensink2014costa} is also remarkably easy under our framework which models class-conditional distributions using exponential family distributions with conjugate priors.

%Also cite and discuss the revealing distribution paper.
% 
% Another problem is the sole reliance on the attributes (our method instead learns a distribution that is only guided by the attributes but does not depend entirely on them).
% 
% Another limitation of these methods is that they lack a proper generative model of the data.
% 
% 
% A fully generative approach offers several advantages: easy to incorporate unlabeled data from both seen as well as unseen classes. Note that ZSL problems, unlabeled data may be very important since we may have very few labelled examples even for the seen classes.
% 
% Some other relevant work: There is work on learning distributions for each class~\cite{mukherjee2016gaussian}. It addresses the issue of intra-class variability by assuming each class to be defined by a distribution.

\section{A Generative Framework For ZSL}
%\vspace{0.5em}
In zero-shot learning (ZSL) we assume there is a total of $S$ seen classes and $U$ unseen classes. Labelled training examples are only available for the seen classes. The test data is usually assumed to come only from the unseen classes, although in our experiments, we will also evaluate our model for the setting where the test data could come from both seen and unseen classes, a setting known as generalised zero-shot learning~\cite{chao2016empirical}. 

We take a generative modeling approach to the ZSL problem and model the class-conditional distribution for an observation $\xv$ from a seen/unseen class $c$ ($c=1,\ldots,S+U$) using an exponential family distribution~\cite{brown1986fundamentals} with natural parameters $\bsym{\theta}_c$
%\vspace{-0.5em}
\beq
  p(\xv|\boldsymbol{\theta}_c) = h(\xv)\exp\left(\boldsymbol{\theta}_c^\top \phi(\xv) - A(\boldsymbol{\theta}_c)\right)
 % \vspace{-0.5em}
\eeq
where $\phi(\xv)$ denotes the sufficient statistics and $A(\bsym{\theta}_c)$ denotes the log-partition function.
We also assume that the distribution parameters $\boldsymbol{\theta}_c$ are given conjugate priors 

\beq 
p(\boldsymbol{\theta}_c|\boldsymbol{\tau}_0,\boldsymbol{\nu}_0) \propto \exp(\boldsymbol{\theta}_c^\top \boldsymbol{\tau}_0 -\boldsymbol{\nu}_0A(\bsym{\theta}_c))
\eeq

Given a test example $\xv_*$, its class $y_*$ can be predicted by finding the class under which $\xv_*$ is most likely (i.e., $y_* = \arg \max_{c} p(\xv_*|\theta_c)$), or finding the class that has the largest posterior probability given $\xv_*$ (i.e., $y_* = \arg \max_{c} p(\theta_c|\xv_*)$). However, doing this requires first estimating the parameters $\{\bsym{\theta}_c\}_{c=S+1}^{S+U}$ of all the unseen classes. 

Given labelled training data from any class modelled as an exponential family distribution, it is straightforward to estimate the model parameters $\bsym{\theta}_c$ using maximum likelihood estimation (MLE), maximum-a-posteriori (MAP) estimation, or using fully Bayesian inference~\cite{murphy2012machine}. However, since there are no labelled training examples from the unseen classes, we cannot estimate the parameters $\{\bsym{\theta}_c\}_{c=S+1}^{S+U}$ of the class-conditional distributions of the unseen classes. 

To address this issue, we learn a model that allows us to predict the parameters $\bsym{\theta}_c$ for any class $c$ using the \emph{attribute vector} of that class via a \emph{gating} scheme, which is basically defined as a linear/nonlinear regression model from the attribute vector to the parameters. As is the common practice in ZSL, the attribute vector of each class may be derived from a human-provide description of the class or may be obtained from an external source such as Wikipedia in form of word-embedding of each class. We assume that the class-attribute of each class is a vector of size $K$. The class-attribute of all the classes are denoted as $\{\av_c\}_{c=1}^{S+U}$, $\av_c \in \mathbb{R}^K$. %parameterize the parameters of both seen and unseen class distributions using the respective class attribute vector via a gating mechanism, which basically means modelling these parameters as functions of the class attributes.

\subsection{Gating via Class-Attributes} 

We assume a regression model from the class-attribute vector $\av_c$ to the parameters $\bsym{\theta}_c$ of each class $c$. In particular, we assume that the class-attribute vector $\av_c$ is mapped via a function $f$ to generate the parameters $\bsym{\theta}_c$ of the class-conditional distribution of class $c$, as follows % the parameters $\boldsymbol{\theta}_c$ of each seen/unseen class $c$ as a function of the class attribute vector $\av_c$, i.e., 
\beq
  \bsym{\theta}_c = f_\theta(\av_c)
\eeq
Note that the function $f_\theta$ itself could consist of multiple functions if $\bsym{\theta}_c$ consists of multiple parameters. For concereteness, and also to simplify the rest of the exposition, we will focus on the case when the class-conditional distribution is a $D$ dimensional Gaussian, for which $\bsym{\theta}_c$ is defined by the mean vector $\bsym{\mu}_c \in \mathbb{R}^D$ and a p.s.d. covariance matrix $\bsym{\Sigma}_c \in \Scal_+^{D\times D}$. Further, we will assume $\bsym{\Sigma}_c$ to be a diagonal matrix defined as $\bsym{\Sigma}_c = \text{diag}(\bsym{\sigma}_c^2)$ where $\bsym{\sigma}_c^2 = [\sigma_{c1}^2,\ldots,\sigma_{cD}^2]$. Note that one can also assume a full covariance matrix but it will significantly increase the number of parameters to be estimated. We model $\bsym{\mu}_c$ and $\bsym{\sigma}_c^2$ as functions of the attribute vector $\av_c$
\beqs
  \bsym{\mu}_c &=& f_{\bsym{\mu}}(\av_c) \\
  \bsym{\sigma}_c^2 &=& f_{\bsym{\sigma}^2}(\av_c)   
\eeqs

Note that the above equations define two regression models. The first regression model defined by the function $f_{\bsym{\mu}}$ has $\av_c$ as the input and $\bsym{\mu}_c$ as the output. The second regression model defined by $f_{\bsym{\sigma}^2}$ has $\av_c$ as the input and $\bsym{\sigma}^2$ as the output. The goal is to learn the functions $f_{\bsym{\mu}}$ and $f_{\bsym{\sigma}^2}$ from the available training data. Note that the form of these functions is a modelling choice and can be chosen appropriately. We will consider both linear as well as nonlinear functions.

\subsection{Learning The Regression Functions}
\label{sec:model}
Using the available training data from all the seen classes $c=1,\ldots,S$, we can form empirical estimates of the parameters $\{\hat{\bsym{\mu}}_c,\hat{\bsym{\sigma}}_c^2\}_{c=1}^S$ of respective class-conditional distributions using MLE/MAP estimation. Note that, since our framework is generative, both labeled as well as unlabeled data from the seen classes can be used to form the empirical estimates $\{\hat{\bsym{\mu}}_c,\hat{\bsym{\sigma}}_c^2\}_{c=1}^S$. This makes our estimates of $\{\hat{\bsym{\mu}}_c,\hat{\bsym{\sigma}}_c^2\}_{c=1}^S$ reliable even if each seen class has very small number of labeled examples. Given these estimates for the seen classes
\beqs
  \hat{\bsym{\mu}}_c &=& f_{\bsym{\mu}}(\av_c) \quad \quad \ c=1,\ldots,S\\
  \hat{\bsym{\sigma}}_c^2 &=& f_{\bsym{\sigma}^2}(\av_c)  \quad \quad c=1,\ldots,S 
\eeqs
We can now learn $f_{\bsym{\mu}}$ using ``training'' data $\{\av_c,\hat{\bsym{\mu}}_c\}_{c=1}^S$ and learn $f_{\bsym{\sigma^2}}$ using training data $\{\av_c,\hat{\bsym{\sigma}^2}_c\}_{c=1}^S$. We consider both linear and nonlinear regression models for learning these. % functions. 

%\noindent 

\subsubsection{The Linear Model} 

For the linear model, we assume $\hat{\bsym{\mu}}_c$ and $\hat{\bsym{\sigma}}_c^2$ to be linear functions of the class-attribute vector $\av_c$, defined as
\beqs
  \hat{\bsym{\mu}}_c &=& \Wmat_{\mu} \av_c \quad \quad \ c=1,\ldots,S \\
  \hat{\bsym{\rho}}_c = \log \hat{\bsym{\sigma}}_c^2 &=& \Wmat_{\sigma^2} \av_c \quad \quad c=1,\ldots,S 
\eeqs 
where the regression weights $\Wmat_{\mu} \in \mathbb{R}^{D\times K}$, $\Wmat_{\sigma^2} \in \mathbb{R}^{D\times K}$, and we have re-parameterized $\hat{\bsym{\sigma}}_c^2  \in \mathbb{R}_+^D$ to $\hat{\bsym{\rho}}_c \in \mathbb{R}^D$ as $\hat{\bsym{\rho}}_c = \log \hat{\bsym{\sigma}}_c^2$. 

We use this re-parameterization to map the output space of the second regression model $f_{\bsym{\sigma}^2}$ (defined by $\Wmat_{\sigma^2}$) to real-valued vectors, so that a standard regression model can be applied (note that $\hat{\bsym{\sigma}}_c^2$ is positive-valued vector).

\vspace{0.5em}
\noindent \textbf{Estimating Regression Weights of Linear Model:} We will denote $\Mmat = [\hat{\bsym{\mu}}_1,\ldots,\hat{\bsym{\mu}}_S] \in \mathbb{R}^{D\times S}$, $\Rmat = [\hat{\bsym{\rho}}_1,\ldots,\hat{\bsym{\rho}}_S] \in \mathbb{R}^{D\times S}$, and $\Amat = [\av_1,\ldots,\av_S]  \in \mathbb{R}^{K\times S}$. We can then write the estimation of the regression weights $\Wmat_{\mu}$ as the following problem 
\beq
  \hat{\Wmat}_{\mu} = \arg \min_{\Wmat_{\mu}} ||\Mmat - \Wmat_{\mu} \Amat||_2^2 + \lambda_{\mu} ||\Wmat_{\mu}||_2^2
\eeq 
This is essentially a multi-output regression~\cite{friedman2001elements} problem $\Wmat_\mu: \av_s \mapsto \hat{\bsym{\mu}}_s$ with least squares loss and an $\ell_2$ regularizer. The solution to this problem is given by
\beq 
\label{eq:l1}
\hat{\Wmat}_{\mu} = \Mmat\Amat^\top (\Amat\Amat^\top + \lambda_{\mu}\Imat_K)^{-1}
\eeq

Likewise, we can then write the estimation of the regression weights $\Wmat_{\sigma^2}$ as the following problem 
\beq
  \hat{\Wmat}_{\sigma^2} = \arg \min_{\Wmat_{\sigma^2}} ||\Rmat - \Wmat_{\sigma^2} \Amat||_2^2 + \lambda_{\sigma^2} ||\Wmat_{\sigma^2}||_2^2
\eeq 
The solution of the above problem is given by
\beq 
\label{eq:l2}
\hat{\Wmat}_{\sigma^2} = \Rmat\Amat^\top (\Amat\Amat^\top + \lambda_{\sigma^2}\Imat_K)^{-1}
\eeq

Given $\hat{\Wmat}_{\mu}$ and $\hat{\Wmat}_{\sigma^2}$, parameters of the class-conditional distribution of each unseen class $c=S+1,\ldots,S+U$ can be easily computed as follows
\vspace{-0.5em}
\beqs
  \hat{\bsym{\mu}}_c &=& \hat{\Wmat}_{\mu} \av_c \\
  \hat{\bsym{\sigma}}_c^2 &=& \exp(\hat{\bsym{\rho}}_c) = \exp(\hat{\Wmat}_{\sigma^2} \av_c)
  \vspace{-0.5em}
\eeqs

\subsubsection{The Nonlinear Model} 

For the nonlinear case, we assume that the inputs $\{\av_c\}_{c=1}^S$ are mapped to a kernel induced space via a kernel function $k$ with an associated nonlinear mapping $\phi$. In this case, using the representer theorem~\cite{scholkopf2001learning}, the solution for the two regression models $f_{\bsym{\mu}}$ and $f_{\bsym{\sigma}^2}$ can be written as the spans of the inputs $\{\phi(\av_c)\}_{c=1}^S$. Note that mappings $\phi(\av_c)$ do not have to be computed explicitly since learning the nonlinear regression model only requires dot products $\phi(\av_c)^\top \phi(\av_{c^\prime}) = k(\av_c,\av_{c^\prime})$ between the nonlinear mappings of two classes $c$ and $c^\prime$. 

\vspace{0.5em}
\noindent \textbf{Estimating Regression Weights of Nonlinear Model:} Denoting $\Kmat$ to be the $S\times S$ kernel matrix of the pairwise similarities of the attributes of the seen classes, the nonlinear model $f_{\bsym{\mu}}$ is obtained by   
\beq
  \hat{\bsym{\alpha}}_{\mu} = \arg \min_{\bsym{\alpha}_{\mu}} ||\Mmat - \bsym{\alpha}_{\mu} \Kmat||_2^2 + \lambda_{\mu} ||\bsym{\alpha}_{\mu} ||_2^2
  \label{eq:kerreg}
\eeq 
where $\hat{\bsym{\alpha}}_{\mu}$ is a $D\times S$ matrix consists of the coefficients of the span of $\{\phi(\av_c)\}_{c=1}^S$ defining the nonlinear function $f_{\bsym{\mu}}$. 

Note that the problem in Equation~\ref{eq:kerreg} is essentially a multi-output \emph{kernel} ridge regression~\cite{friedman2001elements} problem, which has a closed form solution. The solution for $\hat{\bsym{\alpha}}_{\mu}$ is given by
\beq 
\label{eq:nl1}
  \hat{\bsym{\alpha}}_{\mu} = \Mmat(\Kmat + \lambda_{\mu}\Imat_S)^{-1} 
\eeq 

Likewise, the nonlinear model $f_{\bsym{\sigma}^2}$ is obtained by solving    
\beq
  \hat{\bsym{\alpha}}_{\sigma^2} = \arg \min_{\bsym{\alpha}_{\sigma^2}} ||\Mmat - \bsym{\alpha}_{\sigma^2} \Kmat||_2^2 + \lambda_{\sigma^2} ||\bsym{\alpha}_{\sigma^2} ||_2^2
\eeq 
where $\hat{\bsym{\alpha}}_{\sigma^2}$ is a $D\times S$ matrix consists of the coefficients of the span of $\{\phi(\av_c)\}_{c=1}^S$ defining the nonlinear function $f_{\bsym{\sigma^2}}$. The solution for $\hat{\bsym{\alpha}}_{\sigma^2}$ is given by
\beq 
\label{eq:nl2}
  \hat{\bsym{\alpha}}_{\sigma^2} = \Rmat(\Kmat + \lambda_{\mu}\Imat_S)^{-1} 
\eeq 

Given $\hat{\bsym{\alpha}}_{\mu}$, $\hat{\bsym{\alpha}}_{\sigma^2}$,  parameters of  class-conditional distribution of each unseen class $c=S+1,\ldots,S+U$ will be
\beqs
  \hat{\bsym{\mu}}_c &=& \hat{\bsym{\alpha}}_{\mu} \kv_c \\
  \hat{\bsym{\sigma}}_c^2 &=& \exp(\hat{\bsym{\rho}}_c) = \exp(\hat{\bsym{\alpha}}_{\sigma^2}\kv_c)
\eeqs
where $\kv_c = [k(\av_c,\av_1),\ldots,k(\av_c,\av_S)]^\top$ denotes an $S\times 1$ vector of kernel-based similarities of the class-attribute of unseen class $c$ with the class-attributes of all the seen classes.

\subsubsection{Other Exponential Family Distributions}

Although we illustrated our framework taking the example of Gaussian class-conditional distributions, our framework readily generalizes to the case when these distributions are modelled using any exponential family distribution. The estimation problems can be solved in a similar way as the Gaussian case with the basic recipe remaining the same: Form empirical estimates of the parameters $\bsym{\Theta} = \{\hat{\bsym{\theta}}_c\}_{c=1}^S$ for the seen classes using all the available seen class data and then learn a linear/nonlinear regression model from the class-attributes $\Amat$ (or their kernel representation $\Kmat$ in the nonlinear case) to $\bsym{\Theta}$. 

\begin{figure}[!htbp]
\begin{center}
\includegraphics[scale=0.3]{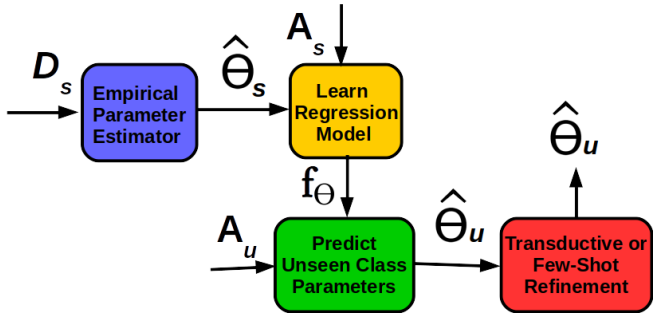}
\caption{\small{Block-diagram of our framework. $\Dcal_s$ denotes the seen class data (can be labeled (and optionally also unlabeled); $\Amat_s$ denotes seen class attributes; $\Amat_u$ denotes unseen class attributes; $\bsym{\hat{\Theta}}_s$ denotes the estimated seen class parameters; $\bsym{\hat{\Theta}}_u$ denotes the estimated unseen class parameters. The last stage - transductive/few-shot refinement - is optional (Section~\ref{sec:tzsl} and \ref{sec:fsl})}}
\label{fig:block-diag}
\end{center}
\end{figure}

In additional to its modeling flexibility, an especially remarkable aspect of our generative framework is that it is very easy to implement, since both the linear model as well as the nonlinear model have closed-form solutions given by Eq.~\ref{eq:l1} and Eq.~\ref{eq:l2}, and Eq.~\ref{eq:nl1} and Eq.~\ref{eq:nl2}, respectively (the solutions will be available in similar closed-forms in the case of other exponential family distributions). A block-diagram describing our framework is shown in Figure~\ref{fig:block-diag}. Note that another appealing aspect of our framework is its modular architecture where each of the blocks in Figure~\ref{fig:block-diag} can make use of a suitable method of one's choice. 

\vspace{-2em}
\subsection{Transductive/Semi-Supervised Setting}
\label{sec:tzsl}
The procedure described in Section~\ref{sec:model} relies only on the seen class data (labeled and, optionally, also unlabeled). As we saw for the Gaussian case, the seen class data is used to form empirical estimates of the parameters $\{\hat{\bsym{\mu}}_c,\hat{\bsym{\sigma}}_c^2\}_{c=1}^S$ of the class-conditional distributions of seen classes, and then these estimates are used to learn the linear/nonlinear regression functions $f_{\bsym{\mu}}$ and $f_{\bsym{\sigma}^2}$. These functions are finally used to compute the parameters $\{\hat{\bsym{\mu}}_c,\hat{\bsym{\sigma}}_c^2\}_{c=S+1}^{S+U}$ of class-conditionals of unseen classes. We call this setting the \emph{inductive} setting. Note that this procedure does not make use of any data from the unseen classes. Sometimes, we may have access to unlabeled data from the unseen classes. 

Our generative framework makes it easy to leverage such \emph{unlabeled} data from the \emph{unseen} classes to further improve upon the estimates $\{\hat{\bsym{\mu}}_c,\hat{\bsym{\sigma}}_c^2\}_{c=S+1}^{S+U}$ of their class-conditional distributions. In our framework, this can be done in two settings, \emph{transductive} and \emph{semi-supervised}, both of which leverage unlabeled data from unseen classes, but in slightly different ways.  If the unlabeled data is the unseen class test data itself, we call it the \emph{transductive} setting. If this unlabeled data from the unseen classes is different from the actual unseen class test data, we call it the \emph{semi-supervised} setting. 

In either setting, we can use an Expectation-Maximization (EM) based procedure that alternates between inferring the labels of unlabeled examples of unseen classes and using the inferred labels to update the estimates of the parameters $\{\hat{\bsym{\mu}}_c,\hat{\bsym{\sigma}}_c^2\}_{c=S+1}^{S+U}$ of the distributions of unseen classes. 

For the case when each class-conditional distribution is a Gaussian, this procedure is equivalent to estimating a Gaussian Mixture Model (GMM) using the unlabeled data $\{\xv_n\}_{n=1}^{N_u}$ from the unseen classes. The GMM is initialized using the estimates $\{\hat{\bsym{\mu}}_c,\hat{\bsym{\sigma}}_c^2\}_{c=S+1}^{S+U}$ obtained from the inductive procedure of Section~\ref{sec:model}. Note that each of the $U$ mixture components of this GMM corresonds to an unseen class.
\vspace{0.5em}\\
The EM algorithm for the Gaussian case is summarized next
\begin{enumerate}
 \item Initialize mixing proportions $\bsym{\pi} = [\pi_1,\ldots,\pi_U]$ uniformly set mixture parameters as $\bsym{\Theta} = \{\hat{\bsym{\mu}}_c,\hat{\bsym{\sigma}}_c^2\}_{c=S+1}^{S+U}$
 \vspace{0.5em}
 \item \textbf{E Step:} Infer the probabilities for each $\xv_n$ belonging to each of the unseen classes $c=S+1,\ldots,S+U$ as
 \vspace{-0.5em}
 \[
  p(y_n=c|\xv_n,\pi,\bsym{\Theta}) = \frac{\pi_c \Ncal(\xv_n|\hat{\bsym{\mu}}_c,\hat{\bsym{\sigma}}_c^2)}{\sum_{c}\pi_c \Ncal(\xv_n|\hat{\bsym{\mu}}_c,\hat{\bsym{\sigma}}_c^2)} % \quad \forall c=S+1,\ldots,S+U
  \vspace{-0.5em}
 \]
 \item \textbf{M Step:} Use to inferred class labels to re-estimate $\bsym{\pi}$ and $\bsym{\Theta} = \{\hat{\bsym{\mu}}_c,\hat{\bsym{\sigma}}_c^2\}_{c=S+1}^{S+U}$. 
 \vspace{0.5em}
 \item Go to step 2 if not converged. 
\end{enumerate}
Note that the same procedure can be applied even when each class-conditional distribution is some exponential family distribution other than Gaussian. The E and M steps in the resulting mixture model are straightforward in that case as well. The E step will simply require the Gausian likelihood to be replaced by the corresponding exponential family distribution's likelihood. The M step will require doing MLE of the exponential family distribution's parameters, which has closed-form solutions.

\vspace{-0.5em}
\subsection{Extension for Few-Shot Learning}
\label{sec:fsl}
In few-shot learning, we assume that a very small number of labeled examples may also be available for the unseen classes~\cite{salakhutdinov2013learning,mensink2014costa}. The generative aspect of our framework, along with the fact the the data distribution is an exponential family distribution with a conjugate prior on its parameters, makes it very convenient for our model to be extended to this setting. The outputs $\{\hat{\bsym{\mu}}_c,\hat{\bsym{\sigma}}_c^2\}_{c=S+1}^{S+U}$ of our generative zero-shot learning model can naturally serve as the hyper-parameters of a conjugate prior on parameters of class-conditional distributions of unseen classes, which can then be updated given a small number of labeled examples from the unseen classes. For example, in the Gaussian case, due to conjugacy, we are able to update the estimates $\{\hat{\bsym{\mu}}_c,\hat{\bsym{\sigma}}_c^2\}_{c=S+1}^{S+U}$ in a straightforward manner when provided with such labeled data. In particular, given a small number of labeled examples $\{\xv_n\}_{n=1}^{N_c}$ from an unseen class $c$, $\hat{\bsym{\mu}}_c$ and $\hat{\bsym{\sigma}}_c^2$ can be easily updated as
\beqs 
  \bsym{\mu}_c^{(FS)} &=& \frac{\hat{\bsym{\mu}}_c + \sum_{n=1}^{N_c} \xv_n}{1+N_c} \\
  {\bsym{\sigma}_c^2}^{(FS)} &=& \left(\frac{1}{\hat{\bsym{\sigma}}_c^2} + \frac{N_c}{\sigma^2}\right)^{-1} 
\eeqs 
where $\sigma^2 = \frac{1}{N_c}\sum_{n=1}^{N_c}(\xv_n -\hat{\bsym{\mu}}_c)^2$ denotes the empirical variance of the $N_c$ observations from the unseen class $c$.

A particularly appealing aspect of our few-shot learning model outlined above is that it can also be updated in an online manner as more and more labelled examples become available from the unseen classes, without having to re-train the model from scratch using all the data.

%If the data is assumed to be generated from some other exponential family distribution, the GMM can be replaced by the appropriate exponential family mixture model. iteratively estimating the cluster memberships of the unlabeled examples  $\{\xv_n\}_{n=1}^{N_u}$, and using these inferred cluster memberships to refine the estimates $\{\hat{\bsym{\mu}}_c,\hat{\bsym{\sigma}}_c^2\}_{c=S+1}^{S+U}$.

% The solution of the above problem is given by
% \beq 
% \hat{\bsym{\alpha}}_{\mu} = \Mmat\Amat^\top (\Amat\Amat^\top + \lambda_{\mu}\Imat_K)^{-1}
% \eeq
% 
% 
% $\hat{\bsym{\mu}}_c = \Wmat_{\mu} \phi(\av_c)$ and $\hat{\bsym{\rho}}_c = \Wmat_{\sigma^2}\phi(\av_c)$. Note that $\Wmat_{\mu}$ and $\Wmat_{\sigma^2}$ are given by the span of the inputs $\{\phi(\av_c)\}_{c=1}^S$. In particular, column  $\Wmat_{\mu} = \sum_{c=1}^S \alpha_c \phi(\av_c)$ and $\Wmat_{\sigma^2} = \sum_{c=1}^S \alpha_c \phi(\av_c)$. 

%Note that usually each $\bsym{\theta}_c$ itself consists of a set of parameters (e.g., for a Gaussian distribution $\bsym{\theta}_c$ would consist of mean and variance). Assuming $\theta_c = \{\theta_{c1},\ldots,\theta_{c\ell}\}$, we model each $\theta_{c\ell}$ using a separate function $f_\ell$ as $\theta_{c\ell} = f_\ell(\av_c)$.

\vspace{-0.5em}
\section{Related Work}

Some of the earliest works on ZSL are based on predicting attributes for each example~\cite{lampert2009learning}. This was followed by a related line of work based on models that assume that the data from each class can be mapped to the class-attribute space (a shared semantic space) in which each seen/unseen class is also represented as a point~\cite{socher2013zero,akata2013label,zhang2015zero56}. The mapping can be learned using various ways, such as linear models or feed forward neural networks or convolutional neural networks. Predicting the label for a novel unseen class example then involves mapping it to this space and finding the ``closest'' unseen class. Some of the work on ZSL is aimed at improving the semantic embeddings of concepts/classes. For example, ~\cite{wang2016relational} proposed a ZSL model to incorporate relational information about concepts. In another recent work,  \cite{bucher2016improving} proposed a model to improve the semantic embeddings using a metric learning formulation. A complementary line of work to the semantic embedding methods is based on a ``reverse'' mapping, i.e., mapping the class-attribute to the observed feature space~\cite{deep2016learning,revealing-data2016zero}. 

In contrast to such semantic embedding methods that assume that the classes are collapsed onto a single point, our framework offers considerably more flexibility by modelling each class using its own distribution. This makes our model more suitable for capturing the intra-class variability, which the simple point-based embedding models are incapable of handling.

Another popular approach for ZSL is based on modelling each unseen class as a linear/convex combination of seen classes~\cite{norouzi2013zero} or of a set of ``abstract'' or ``basis'' classes~\cite{romera2015embarrassingly,changpinyo2016synthesized}. The latter class of methods, in particular, can be seen as a special case of our framework since, for our linear model, we can view the columns of the $D\times K$ regression weights as representing a set of $K$ basis classes. Note however that our model has such regression weights for each parameter of the class-conditional distribution, allowing it to be considerably more flexible. Moreover, our framework is also significantly different in other ways due to its fully generative framework, due to its ability to incorporate unlabeled data, performing few-shot learning, and its ability to model different types of data using an appropriate exponential family distribution.

A very important issue in ZSL is the \emph{domain shift} problem which may arise if the seen and unseen class come from very different domains. In these situations, standard ZSL models tend to perform badly. This can be somewhat alleviated using some additional unlabeled data from the unseen classes. To this end, \cite{unsupervised2015domain} provide a dictionary learning based approach for learning unseen class classifiers in which the dictionary is adapted to the unseen class domain. The dictionary adaptation is facilitated using unlabeled data from the unseen classes. In another related work, \cite{fu2015transductive} leverage unlabeled data in a transductive ZSL framework to handle the domain shift problem. Note that our framework is robust to the domain shift problem due to its ability to incorporate unlabeled data from the unseen classes (the transductive setting). Our experimental results corroborate this.

Semi-supervised learning for ZSL can also be used to improve the semantic embedding based methods. \cite{li2015semi} provide a semi-supervised method that leverages prior knowledge for improving the learned embeddings. In another recent work, \cite{revealing-data2016zero} present a model to incorporate unlabeled unseen class data in a setting where each unseen class is represented as a linear combination of seen classes.  \cite{zhang2016learningDec} provide another approach, motivated by applications in computer vision, that jointly facilitates the domain adaptation of attribute space and the visual space. Another semi-supervised approach presented in~\cite{li2015max} combines a semisupervised classification model over the observed classes with an unsupervised clustering model over unseen classes together to address the zero-shot multi-class classification.

In contrast to these models for which the mechanism for incorporating unlabeled data is model-specific, our framework offers a general approach for doing this, while also being simple to implement. Moreover, for large-scale problems, it can also leverage more efficient solvers (e.g., gradient methods) for estimating the regression coefficients associated with class-conditional distributions. % for large-scale problems. %  we are providing a generative framework that predicts the unseen class parameter based on the attribute and our approach are closely related to \cite{revealing-data2016zero}.

\vspace{-1em}
\section{Experiments}

We evaluate our generative framework for zero-shot learning (hereafter referred to as {\bf GFZSL}) on several benchmark data sets and compare it with a number of state-of-the-art baselines. We conduct our experiments on various problem settings, including standard \emph{inductive} zero-shot learning (only using seen class labeled examples), \emph{transductive} zero-shot learning (using seen class labeled examples and unseen class unlabeled examples), and few-shot learning (using seen class labeled examples and a very small number of unseen class labeled examples). We report our experimental results on the following benchmark data sets:

\begin{itemize}
 \item \textbf{Animal with Attribute({AwA})}: The AwA data set contains 30475 images with 40 seen classes (training set) and 10 unseen classes (test set). Each class has a human-provided binary/continuous 85-dimensional class-attribute vector~\cite{krizhevsky2009learningAwA}. We use continuous class-attributes since prior works have found these to have more discriminative power.
 \item \textbf{Caltech-UCSD Birds-200-2011 ({CUB-200}):} The CUB-200 data set contains 11788 images with 150 seen classes (training set) and 50 unseen class (test set). Each image has a binary 312-dimensional class-attribute vector, specifying the presence or absence of various attribute of that image~\cite{wah2011caltechCUB}. The attribute vectors for all images in a class are averaged to construct its continuous class-attribute vector~\cite{akata2015evaluation}.  We use the same train/test split for this data set as used in \cite{akata2015evaluation}.
 \item \textbf{SUN attribute (SUN):} The SUN data set contains 14340 images with 707 seen classes (training set) and 10 unseen classes (test set). Each image is described by a 102-dimensional binary class-attribute vector. Just like the CUB-200 data set, we average the attribute vectors of all images in each class to get its continuous attribute vector~\cite{jayaraman2014zeroSUNsplit}. We use the same train/test split for this data set as used in \cite{jayaraman2014zeroSUNsplit}. % and other methods.
\end{itemize}
\vspace{-0.5em}
For image features, we considered both GoogleNet features~\cite{googlenet2015going} and VGG-19(4096) fc7 features~\cite{simonyan2014veryVGG-19} and found that our approach works better with VGG-19. All of the state-of-the-art baselines we compare with in our experiments use  VGG-19 fc7 features or GoogleNet features~\cite{googlenet2015going}. For the nonlinear (kernel) variant of our model, we use a quadratic kernel. Our set of experiments include:

\begin{itemize}
 \item \textbf{Zero-Shot Learning:} We consider both inductive ZSL as well as transductive ZSL.
 \begin{itemize}
  \item \textbf{Inductive ZSL:} This is the standard ZSL setting where the unseen class parameters are learned using only seen class data.
  \item \textbf{Transductive  ZSL:} In this setting~\cite{zhang2016learningDec}, we also use the unlabeled test data while learning the unseen class parameters. Note that this setting has access to more information about the unseen class; however, it is only through unlabeled data.
 \end{itemize}

\noindent 
%  \begin{itemize}
%  \item \textbf{Inductive ZSL:} This is the standard ZSL setting where the unseen class parameters are learned using only seen class data.
%  \item \textbf{Transductive  ZSL:} In this setting~\cite{zhang2016learningDec}, we also use the unlabeled test data while learning the unseen class parameters. Note that this setting has access to more information about the unseen class, even though it is only from unlabeled data.
%  \end{itemize}
 \item \textbf{Few-Shot Learning:} In this setting~\cite{salakhutdinov2013learning,mensink2014costa}, we also use a small number of labelled examples from each unseen class.
 \item \textbf{Generalized ZSL:} Whereas standard ZSL (as well as few-shot learning) assumes that the test data can only be from the unseen classes, generalized ZSL assumes that the test data can be from unseen as well as seen classes. This is usually a more challenging setting~\cite{chao2016empirical} and most of the existing methods are known to be biased towards predicting the seen classes.
\end{itemize}

%\subsection{Model Setting and Training}
% \subsubsection*{Inductive \& Transductive Setting} 
% In the inductive setting we assume that for the test time the only single data point is available. This is the extreme case of the streaming data. In the transductive setting, we assume that all data are present at the test time. So transductive setting is more informative than inductive, Because we can infer here more information about the data like distribution, the similarity between data point etc.
%\subsubsection*{Model Selection} 
We use the standard train/test split as given in the data description section. For selecting the hyperparameters, we further divide the train set further into train and validation set. In our model, we have two hyper-parameter $\lambda_{\mu}$ and $\lambda_{\sigma^2}$, which we tune using the validation dataset. For AwA, from the 40 seen classes, a random selection of 30 classes are used for the training set and 10 classes are used for the validation set. For CUB-200, from the 150 seen classes, 100 are used for the training set and rest 50 are used for the validation set. Similarly, for the SUN dataset from the 707 seen classes, 697 are used for the training set and rest 10  is used for the validation set. We use cross-validation on the validation set to choose the best hyperparameter  $[\lambda_{\mu}, \lambda_{\sigma^2}]$ for the each data set and use these for testing on the unseen classes. % After selecting the best hyperparameter we combine the train and validation set and retrain for getting the [$\mu*,\Sigma*$].  The same parameters are used for both the transductive and inductive setting.

\subsection{Zero-Shot Learning}

In our first set of experiments, we evaluate our model for zero-shot learning and compare with a number of state-of-the-art methods, for the inductive setting (which uses only the seen class labelled data) as well as the transductive setting (which uses the seen class data and the unseen class unlabeled data).

\subsubsection{Inductive ZSL} Table-\ref{table-inductive} shows our results for the inductive ZSL setting. The results of the various baselines are taken from the corresponding papers. As shown in the Table-\ref{table-inductive}, on CUB-200 and SUN, both of our models (linear and nonlinear) perform better than all of the other state-of-the-art methods. On AwA, our model has only a marginally lower test accuracy as compared to the best performing baseline~\cite{zhang2016learningDec}. However, we also have an average improvement 5.67\% on all the 3 data sets as compared to the overall best baseline~\cite{zhang2016learningDec}. Among baselines using VGG-19 features (bottom half of Table-\ref{table-inductive}), our model achieves a 21.05\% relative improvement over the best baseline on the CUB-200 data, which is considered to be a difficult data set with many fine-grained classes. 

\begin{table}[!htbp]
\scriptsize \addtolength{\tabcolsep}{13.0pt}
\begin{center}
 \begin{tabular}{||l|l|l|l|l||}
 \hline
  Method & AwA & CUB-200 & SUN & Average\\ [0.5ex] 
 \hline
 \hline
%\cite{farhadi2009describingP&Y} & -- & -- & --& 32.5 &-- \\ 
%\cite{jayaraman2014zeroSUNsplit} & 43.01$\pm$ 0.07 & -- & 56.18$\pm$ 0.27&  26.02$\pm$0.05 &-- \\
%\cite{akata2013label} & 43.50 & 18.0 & --&  -- & --\\
%\cite{lampert2014attribute} & 40.50 & --& 52.5 & 19.1 &--\\ 
%\cite{romera2015embarrassingly} &49.30&-- & 65.75$\pm$0.51& 27.27$\pm$1.62 &--\\
 %\hline
 %\hline
 Akata et al. \cite{akata2015evaluation} & 66.70& 50.1&--&--\\
 Qiao et al. \cite{qiao2016less} & 66.46$\pm$0.42& 29$\pm$0.28&--&--\\
 Xian et al. \cite{xian2016latent}& 71.9 & 45.5 &-- &--\\
 Changpimyo et al.\cite{changpinyo2016synthesized} & 72.9 & 54.7 & 62.7 & 63.43\\
Wang et al.\cite{wang2016relational} & 75.99 & 33.48 &-- &--\\
 \hline \hline
 Lampert et al.\cite{lampert2014attribute} & 57.23&--& 72.00& --\\
 Romera and Torr\cite{romera2015embarrassingly} & 75.32$\pm$2.28& --& 82.10$\pm$ 0.32 & --\\
 Bucher et al.\cite{bucher2016improving} & 77.32$\pm$1.03 & 43.29$\pm$0.38 & 84.41$\pm$0.71&  68.34\\
 %\cite{zhang2015zero} &&&&\\
 Z. Zhang et al.\cite{zhang2016zero57} & 79.12$\pm$0.53 & 41.78$\pm$0.52 & 83.83$\pm$.29 & 68.24\\
 Wang et al.\cite{wang2016zero} &  79.2$\pm$0.0 & 46.7$\pm$0.0 &--&--\\
 Z. Zhang et al.\cite{zhang2016learningDec} &{\bf 81.03$\pm$0.88} & 46.48$\pm$1.67 & 84.10$\pm$1.51& 70.53\\
 {\bf GFZSL: Linear} & 79.90&  52.09 &  86.50&  72.23\\
{\bf GFZSL: Nonlinear} & 80.83 & {\bf 56.53} & {\bf 86.50} & {\bf 74.59}\\
\hline
\end{tabular}
\vspace{1.5em}
\caption{\small{Accuracy(\%) of different type of images features. {\bf Top:} Deep features like AlexNet, GoogleNet, etc. {\bf Bottom:} Deep VGG-19 features. The '-' indicates that this result was not reported.}} %\vspace{-0.5em}
\label{table-inductive}
\end{center}
\vspace{-3em}
\end{table}

In contrast to other models that embed the test examples in the semantic space and then find the most similar class by doing a Euclidean distance based nearest neighbor search, or models that are based on computing the similarity scores between seen and unseen classes~\cite{zhang2015zero56}, for our models, finding the ``most probable class'' corresponds to computing the distance of each test example from a \emph{distribution}. This naturally takes into account the shape and spread of the class-conditional distribution. This explains the favourable performance of our model as compared to the other methods.

%Our simplest method has very stable result compare to all other existing methods that has a significant standard deviation. Here we are estimating the visual embedding based on the attribute vector and instead of searching the nearest neighbour based on Euclidean distance we are using the manifold distance(Gaussian). The CUB-200 dataset is fine-grain dataset and has 200 bird class. This dataset is very challenging dataset and to classify only based on the attribute is very difficult task, but our approach shows the excellent improvement on this data.
\subsubsection{Transductive Setting} For transductive ZSL setting~\cite{guo2016transductive,zhang2016zero57,zhang2016zerostruct58}, we follow the procedure described in Section~\ref{sec:tzsl} to estimate parameters of the class-conditional distribution of each unseen class. After learning the parameters, we find the most probable class for each test example by evaluating its probability under each unseen class distribution and assign it to the class under which it has the largest probability. Table-\ref{table-transductive1} and \ref{table-transductive2} compare our results from the transductive setting with other state-of-the-art baselines designed for the transductive setting. In addition to accuracy, we also report precision and recall results of our model and the other baselines (wherever available). As we can see from Table-\ref{table-transductive1} and \ref{table-transductive2}, both of our models (linear and kernel) outperform the other baselines on all the 3 data sets. Also comparing with the inductive setting results presented in Table-\ref{table-inductive}, we observe that our generative framework is able to very effectively leverage unlabeled data and significantly improve upon the results of a purely inductive setting 

\begin{table*}[!htbp]
\small
\addtolength{\tabcolsep}{3.0pt}
\begin{center}
 \begin{tabular}{||l|l|l|l|l||}
 \hline
  Method & AwA & CUB-200 & SUN & Average \\ [0.5ex] 
 \hline
 \hline
Guo et al.\cite{guo2016transductive} & 78.47 & -- & 82.00 & -- \\ 
Romera et al.\cite{romera2015embarrassingly}+ Zhang et al. \cite{zhang2016zerostruct58} & 84.30&-- & 37.50&-- \\
Zhang et al.\cite{zhang2016zero57}+Zhang et al. \cite{zhang2016zerostruct58} & 92.08$\pm$0.14& 55.34$\pm$0.77& 86.12$\pm$0.99& 77.85  \\
Zhang et al.\cite{zhang2016learningDec}+Zhang et al. \cite{zhang2016zerostruct58} & 88.04$\pm$0.69& 55.81$\pm$1.37& 85.35$\pm$1.56 & 76.40  \\
%\cite{wang2016zero} &{\bf 95.10$\pm$0}& 58.8$\pm$0.1&--&--&--&--&--&--&--&--\\
{\bf GFZSL: Linear} & {94.20}&  57.14 &  87.00& 79.45\\
{\bf GFZSL: Kernel} & {\bf 94.25}& {\bf 63.66}& {\bf 87.00} & {\bf 80.63} \\
 \hline
\end{tabular}
\vspace{1.5em}
\caption{\small{ZSL accuracy(\%) obtained in the transductive setting: results reported using the VGG-19 feature. Average Precision and recall for the all dataset with its standard daviation over the 100 iteration. The '-' indicates that this result was not reported in the original paper.}}
\label{table-transductive1}
\end{center}
\end{table*}

\begin{table*}[!htbp]
\scriptsize
\addtolength{\tabcolsep}{0.0pt}
\begin{center}
 \begin{tabular}{||l|l|l|l|l|l|l||}
 \hline
  \multicolumn{1}{|c|}{} &\multicolumn{3}{|c|}{{\bf Average Precision}} &\multicolumn{3}{|c|}{{\bf Average Recall}} \\
  \hline
 \hline
  Method & AwA & CUB-200 & SUN & AwA & CUB-200 & SUN\\ [0.5ex] 
 \hline
 \hline
%Guo et al.\cite{guo2016transductive} &--&--&--&--&--&--\\ 
%Romera et al.\cite{romera2015embarrassingly}+ Zhang et al. \cite{zhang2016zerostruct58} &--&--&--&--&--&--\\
Zhang et al.\cite{zhang2016zero57}+Zhang et al. \cite{zhang2016zerostruct58} & 91.37$\pm$14.75 & 57.09$\pm$27.91&  85.96$\pm$10.15 & 90.28$\pm$8.08 & 55.73$\pm$31.80 & 86.00$\pm$13.19 \\
Zhang et al.\cite{zhang2016learningDec}+Zhang et al. \cite{zhang2016zerostruct58}   & 89.19$\pm$15.09 & 57.20$\pm$25.96 & 86.06$\pm$ 12.36 & 86.04$\pm$9.82& 55.77$\pm$26.54 & 85.50$\pm$13.68\\
%\cite{wang2016zero} &{\bf 95.10$\pm$0}& 58.8$\pm$0.1&--&--&--&--&--&--&--&--\\
{\bf GFZSL: Linear}&  93.70 &  57.90  &  87.40 & 92.20 &  57.40  &  87.00\\
{\bf GFZSL: Kernel} & {\bf 93.80} & {\bf 64.09}  & {\bf 87.40} & {\bf 92.30} & {\bf 63.96}  & {\bf 87.00}\\
 \hline
\end{tabular}
\vspace{1.5em}
\caption{\small{ZSL precision and recall scores obtained in the transductive setting: results reported using the VGG-19 features. Average Precision and recall for the all dataset with its standard daviation over the 100 iteration. Note: Precision and recall scores not available for Guo et al.\cite{guo2016transductive} and Romera et al.\cite{romera2015embarrassingly}+ Zhang et al. \cite{zhang2016zerostruct58}}}
\label{table-transductive2}
\end{center}
\vspace{-4em}
\end{table*}

\begin{table}[!htbp]
\scriptsize
\addtolength{\tabcolsep}{6.5pt}
\begin{center}
 \begin{tabular}{||l|l|l|l|l|l|l||}
 \hline
  Dataset & Method & 2 & 5 & 10 & 15 & 20   \\ [0.5ex] 
 \hline
 \hline
\multirow{2}{4em}{AwA} & GFZSL & \bf{87.96$\pm$1.47}& \bf{91.64 $\pm$0.81}& \bf{93.31 $\pm$0.50}& \bf{94.01 $\pm$.36}& \bf{94.30 $\pm$0.33}\\
\cline{2-7}
& SVM & 74.81 & 83.19 & 90.44 & 91.22& 92.04\\
\hline
\multirow{2}{4em}{CUB-200} & GFZSL & 60.84 $\pm$1.39 & 64.81$\pm$ 1.14& 68.44$\pm$ 1.21& 70.11$\pm$ 0.93& 71.23$\pm$ 0.87 \\
\cline{2-7}
& SVM & 46.19& 59.33 & 68.75 & 73.87 & 75.42\\
\hline
\multirow{2}{4em}{SUN} & GFZSL & 75.57$\pm$ 4.79 & 83.05$\pm$ 3.60& 82.09 $\pm$3.30&--&-- \\
\cline{2-7}
& SVM & 56.00& 77.00 & 78.00 & -- &--\\
%aP\&Y &87.53$\pm$ 2.19 & 92.05$\pm$ .91 & 93.72$\pm$ 0.67& 94.52$\pm$ 0.55 & 94.93$\pm$ 0.5\\
\hline
\end{tabular}
\vspace{1.5em}
\caption{\small{Accuracy(\%) in the few-shot learning setting: For each data set, the accuracies are reported using $2,5,10,15,20$ labeled examples for each unseen class}}
\label{Few-shot}
\end{center}
\end{table}

\subsection{Few-shot Learning (FSL)}
\label{sec:fsl}
We next perform an experiment with the few-shot learning setting~\cite{salakhutdinov2013learning,mensink2014costa} where we provide each model with a small number of labelled examples from each of the unseen classes. For this experiment, we follow the procedure described in Section~\ref{sec:fsl} to learn the parameters of the class-conditional distributions of the unseen classes. In particular, we train the inductive ZSL model (using only the seen class training data) and the refine the learned model further using a very small number of labelled examples from the unseen classes (i.e., the few-shot learning setting). 

\begin{figure}[!htbp]
%\vspace{-2em}
\begin{center}
\includegraphics[height=6cm,width=10.0cm]{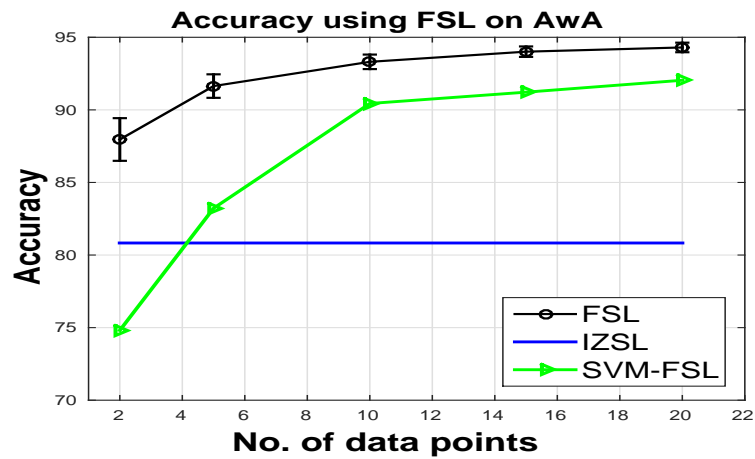}
\caption{\small{(On AwA data): A comparison on classification accuracies of the few-shot learning variant of our model with multi-class SVM (training on labeled examples from seen classes) and the inductive ZSL}}
\label{few-shotfig}
\end{center}
%\vspace{-2em}
\end{figure}

To see the effect of knowledge transfer from the seen classes, we use a multiclass SVM as a baseline that is provided with the same number of labelled examples from each unseen class. In this experiment, we vary the number of labelled examples of unseen classes from 2 to 20 (for SUN we only use 2, 5, and 10 due to the small number of labelled examples). In Figure-\ref{few-shotfig}, we also compare with standard (inductive) ZSL which does not have access to the labelled examples from the unseen classes. Our results are shown Table-\ref{Few-shot} and Figure-\ref{few-shotfig}. 

As shown in Table-\ref{Few-shot} (all data sets) and Figure-\ref{few-shotfig}, the classification accuracy on the unseen classes shows a significant improvement over the standard inductive ZSL, even with as few as 2 or 5 additional labelled examples per class. We also observe that the few-shot learning method outperform multiclass SVM which only relies on the labelled data from the unseen classes. This demonstrates the advantage of the knowledge transfer from the seen class data.

\subsection{Generalized Few-Shot Learning (GFSL)}
We finally perform an experiment on the more challenging generalized few-shot learning setting~\cite{chao2016empirical}. This setting assumes that test examples can come from seen as well as unseen classes. This setting is known to be notoriously hard~\cite{chao2016empirical}. In this setting, although the ZSL models tend to do well on predicting test examples from seen classes, the performance on correctly predicting the unseen class example is poor~\cite{chao2016empirical} since the trained models are heavily biased towards predicting the seen classes. 

One way to mitigate this issue could be to use some labelled examples from the unseen classes (akin to what is done in few-shot learning). We, therefore, perform a similar experiment as in Section~\ref{sec:fsl}. In Table-\ref{genFew-shot}, we show the results of our model on classifying the unseen class test examples in this setting. 

As shown in Table-\ref{genFew-shot}, our model's accuracies on the generalized FSL task improve as it gets to see labelled examples from unseen classes. However, it is still outperformed by a standard multiclass SVM. The better performance of SVM can be attributed to the fact that it is not biased towards the seen classes since the classifier for each class (seen/unseen) is learned independently. 

Our findings are also corroborated by other recent work on generalized FSL~\cite{chao2016empirical} and suggest the need of finding more robust ways to handle this setting. We leave this direction of investigation as a possible future work. 

\begin{table}[!htbp]
\vspace{-1em}
\scriptsize
\addtolength{\tabcolsep}{5.5pt}
\begin{center}
 \begin{tabular}{||l|l|l|l|l|l|l||}
 \hline
  Dataset & Method & 2 & 5 & 10 & 15 & 20 \\ [0.5ex] 
 \hline
\multirow{2}{4em}{AwA} & GFZSL & 25.32 $\pm$   2.43 & 37.42  $\pm$  1.60 & 43.20  $\pm$  1.39
 & 45.09 $\pm$   1.17 & 45.96  $\pm$  1.09 \\
 \cline{2-7}
 & SVM & 40.84 & 60.81 & 75.36 & 77.00 & 77.10\\
 \hline
\multirow{2}{4em}{CUB-200} & GFZSL & 6.64  $\pm$  0.87 & 15.12  $\pm$  1.17 & 22.02  $\pm$  0.76 & 25.03 $\pm$    0.71 & 26.47 $\pm$   0.83 \\
 \cline{2-7}
& SVM & 25.97 & 37.98 & 47.10 &53.87 & 54.42\\
\hline
\multirow{2}{4em}{SUN} & GFZSL & 1.17 $\pm$   1.16 & 4.20  $\pm$ 1.77 & 9.48 $\pm$  2.22&--&--\\
 \cline{2-7}
& SVM & 9.94 & 20.00 & 27.00 &-- &--\\
\hline
%aP\&Y & 46.23 $\pm$   3.63 & 66.79 $\pm$   1.59 & 76.78 $\pm$   1.08 & 80.29 $\pm$   0.92 & 82.01 $\pm$   0.74 \\s
\end{tabular}
\vspace{1.5em}
\caption{\small{Accuracies (\%) in the generalized few-shot learning setting.}}
\label{genFew-shot}
\end{center}
\vspace{-3em}
\end{table}

% 
% \begin{figure}[!htbp]
% \vspace{-2em}
% \begin{center}
% \includegraphics[height=3.3cm,width=4.0cm]{plot/AwA_gen.eps}
% \caption{\small{(On AwA data): The generalized few-shot learning setting}}
% \label{gen-shotfig}
% \end{center}
% \end{figure}

\section{Conclusion}
We have presented a flexible generative framework for zero-shot learning, which is based on modelling each seen/unseen class using an exponential family class-conditional distribution. In contrast to the semantic embedding based methods for zero-shot learning which model each class as a point in a latent space, our approach models each class as a distribution, where the parameters of each class-conditional distribution are functions of the respective class-attribute vectors. Our generative framework allows learning these functions easily using seen class training data (and optionally leveraging additional unlabeled data from seen/unseen classes).

An especially appealing aspect of our framework is its simplicity and modular architecture (cf., Figure~\ref{fig:block-diag}) which allows using a variety of algorithms for each of its building blocks. As we showed, our generative framework admits natural extensions to other related problems, such as transductive zero-shot learning and few-shot learning. It is particularly easy to implement and scale to a large number of classes, using advances in large-scale regression. Our generative framework can also be extended to jointly learn the class attributes from an external source of data (e.g., by learning an additional embedding model with our original model). This can be an interesting direction of future work. Finally, although we considered a point estimation of the parameters of class-conditional distributions, it is also possible to take a fully Bayesian approach for learning these distributions. We leave this possibility as a direction for future work.   % reduces to solving a set of linear/nonlinear regression problems for 
% \newpage

\textbf{Acknowledgements:} This work is supported by a grant from Tower Research CSR, Dr. Deep Singh and Daljeet Kaur Fellowship, and Research-I Foundation, IIT Kanpur. Vinay Verma acknowledges support from Visvesvaraya Ph.D. fellowship.

\bibliographystyle{acm}
\bibliography{ecml.bib}
\end{document}